\documentclass[letterpaper, 10 pt, conference]{ieeeconf}  % Comment this line out if you need a4paper

\IEEEoverridecommandlockouts                              % This command is only needed if 
\usepackage{subcaption}

\title{\LARGE \bf
Quantitative Outcome-Oriented Assessment \\ of Microsurgical Anastomosis
}

\author{Luyin Hu$^{1}$, Soheil Gholami$^{1}$, George Dindelegan$^{2}$, Torstein R. Meling$^{3}$, and Aude Billard$^{1}$% <-this % stops a space
\thanks{This work was supported by the Swiss National Science Foundation (Grant No. 9824) and was conducted in collaboration with Iuliu Hatieganu University of Medicine and Pharmacy
in Cluj-Napoca, Romania, and Erasmus University Medical Center in Rotterdam, the Netherlands. We would especially like to thank Drs. Victor Volovici, Nicolay Gabrovsky, and Ruben Dammers for their assistance during the data collection process.}% <-this % stops a space
\thanks{$^{1}$Learning Algorithms and Systems Laboratory (LASA), École Polytechnique Fédérale de Lausanne (EPFL), Switzerland. E-mail: {\tt\small luyin.hu@epfl.ch}
        }%
\thanks{$^{2}$Iuliu Hatieganu University of Medicine and Pharmacy,  Cluj-Napoca, Romania. }%
\thanks{$^{3}$Department of Neurosurgery, National Hospital of Denmark, Copenhagen, Denmark.
}
}

\usepackage{graphicx}%
\usepackage{multirow}%
\usepackage{amsmath,amssymb,amsfonts}%
\usepackage{mathrsfs}%
\usepackage{textcomp}%
\usepackage{manyfoot}%
\usepackage{booktabs}%
\usepackage{algorithm}%
\usepackage{algorithmicx}%
\usepackage{algpseudocode}%
\usepackage{listings}%

\usepackage{adjustbox}
\usepackage{tabularx}
\usepackage[flushleft]{threeparttable}
\usepackage[table,xcdraw]{xcolor}

\begin{document}

\maketitle
\thispagestyle{empty}
\pagestyle{empty}

\begin{abstract}
Microsurgical anastomosis demands exceptional dexterity and visuospatial skills, underscoring the importance of comprehensive training and precise outcome assessment. Currently, methods such as the outcome-oriented anastomosis lapse index are used to evaluate this procedure. However, they often rely on subjective judgment, which can introduce biases that affect the reliability and efficiency of the assessment of competence. 
Leveraging three datasets from hospitals with participants at various levels, we introduce a quantitative framework that uses image-processing techniques for objective assessment of microsurgical anastomoses. The approach uses geometric modeling of errors along with a detection and scoring mechanism, enhancing the efficiency and reliability of microsurgical proficiency assessment and advancing training protocols.
The results show that the geometric metrics effectively replicate expert raters' scoring for the errors considered in this work.

\indent \textit{Clinical relevance}— Our proposed framework enables automated and detailed analysis of errors defined by the ALI score method, a widely used approach in microsurgical training programs for evaluating anastomosis skills. 
\end{abstract}

\section{Introduction}
\label{sec:introduction}
Proficiency and skill assessment in surgical education can be conducted using process-oriented, outcome-oriented, or combined approaches. This evaluation can be performed subjectively by a team of expert surgeons or objectively through sensors that collect and analyze data to assess the trainees' skills. 
Process-oriented approaches delve into a trainee's specific steps and techniques to accomplish a surgical intervention, such as an end-to-end anastomosis. During the assessment phase, these approaches typically study hand motions, movement efficiency, and instrument handling. Among subjective methods, notable assessment methodologies include GRS \cite{selber2012tracking}, OSATS \cite{martin1997objective}, O-SCORE \cite{gofton2012ottawa}, and SMaRT scale \cite{satterwhite2014stanford}, the latter being intricately tailored for microsurgical skills evaluation. 
Outcome-oriented approaches, on the other hand, center on evaluating task outcomes. For example, the SMaRT scale and the ALI score \cite{ghanem2016anastomosis} are employed in microsurgery to assess proficiency levels. These subjective methods require a group of highly expert surgeons to grade the tasks and outcomes individually. This process is time-consuming considering the busy schedules of experts and may introduce subjectivity and bias \cite{moorthy2003objective}, challenging the reliability of these approaches.

Objective skill assessment techniques using sensors for a more unbiased evaluation have been proposed in the surgical training and skill assessment literature for both process-oriented and outcome-oriented evaluation. In the former, i.e., process-oriented approaches, these methods involve analyzing temporal metrics and monitoring hand and instrument motions through camera-based tracking \cite{ebina2021motion}, although they have seen fewer applications in microsurgical skills \cite{gholami2024objective}. 
Regarding the latter, i.e., outcome-oriented approaches, which is the focus of this manuscript, Kim et al. \cite{kim2020end} analyzed microsurgical anastomosis photos and provided a scoring system using image processing software. Although this method delineates the overall variance between the current stitch layout and the ideal equidistant placement, it fails to provide constructive feedback to trainees. Specifically, it does not identify which stitch is causing the error, limiting its effectiveness as feedback for medical trainees.

\begin{figure}[!t]
    \centering
    \includegraphics[width=0.425\textwidth]{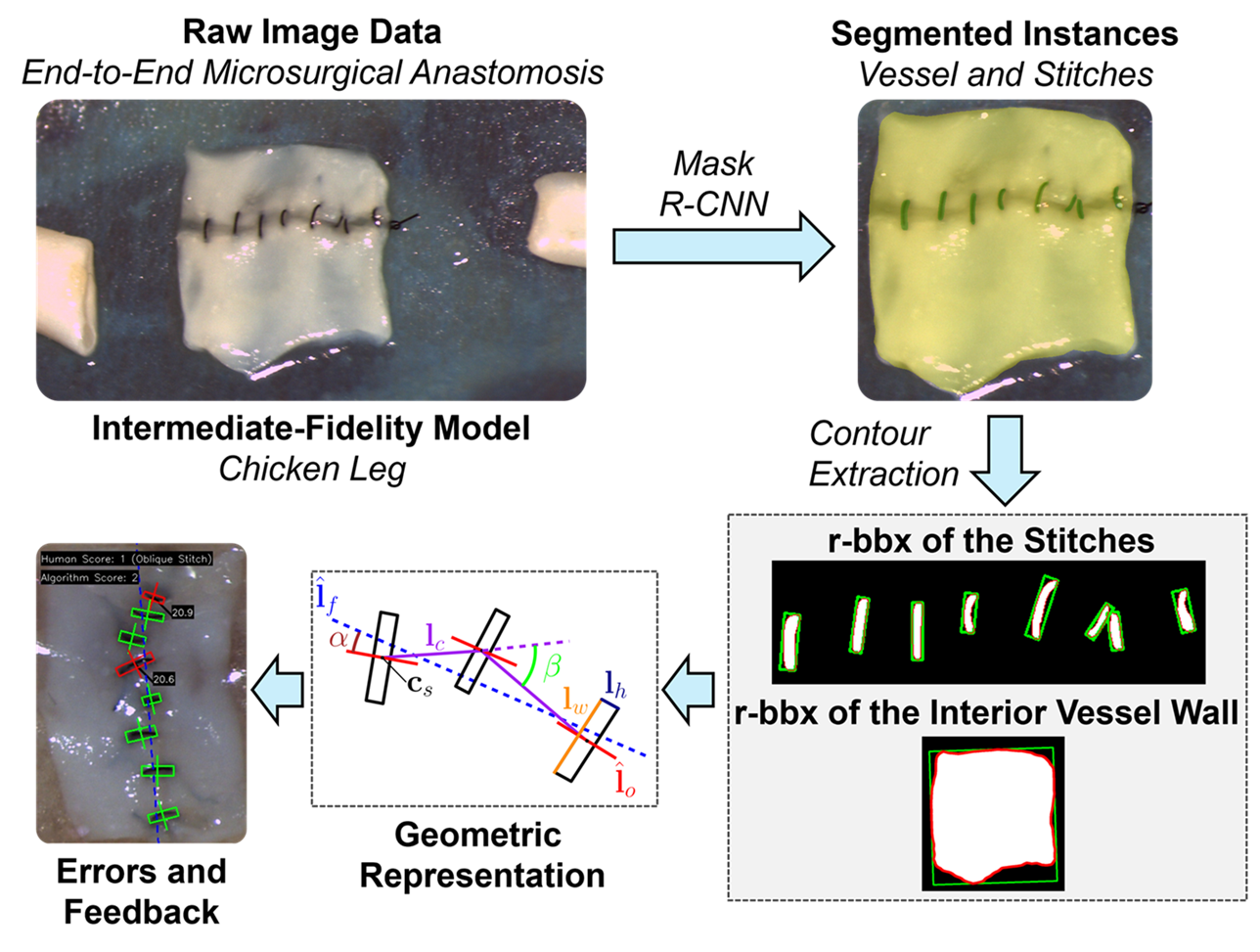} 
    \caption{Proposed framework for objective and quantitative assessment and scoring of microsurgical anastomosis photos.}
    \label{fig:fig1}
\end{figure}

Therefore, a more systematic and quantitative approach to assessing anastomosis photos in microsurgical interventions is essential. As illustrated in Fig.~\ref{fig:fig1}, we propose a quantitative framework to geometrically model and detect errors in anastomosis photos using image processing techniques. To the best of our knowledge, only one study \cite{kim2020end} employs image processing for assessing microsurgical anastomosis. However, this approach depends on software that necessitates manual labeling and lacks automation. Another related study \cite{noraset2024automated}, which focuses on wound suturing rather than microsurgical anastomosis, also exists, but does not address the specific needs of microsurgical procedures.

The \textit{contributions} of this manuscript are as follows:
\begin{itemize}
\item 
We statistically analyze the reliability of the ALI score method,  by examining the ratings from four expert microsurgeons. This analysis reveals large variations across scorers and hence justifies the necessity of a quantitative and objective framework in this context.
\item 
We offer an automatic scoring system for five of the ten ALI scores. 
Compared to the image processing approach offered in \cite{kim2020end}, we develop a fully automated pipeline for extracting ALI scores from raw images. 

\item We use an intermediate-fidelity model (anastomosis of chicken legs), which more closely resembles real human or animal tissues as compared to the low-fidelity models used in \cite{noraset2024automated}. This similarity includes aspects such as deformation, colors, and the presence of blood.

\item Additionally, we provide an example of what systematic feedback for trainees, based on automatic assessment of errors, could look like.
\end{itemize}

\section{Microsurgical Anastomosis}
\label{subsection:microsurgical-anastomosis}

\begin{figure}[!t]
    \centering
\includegraphics[width=0.49\textwidth]{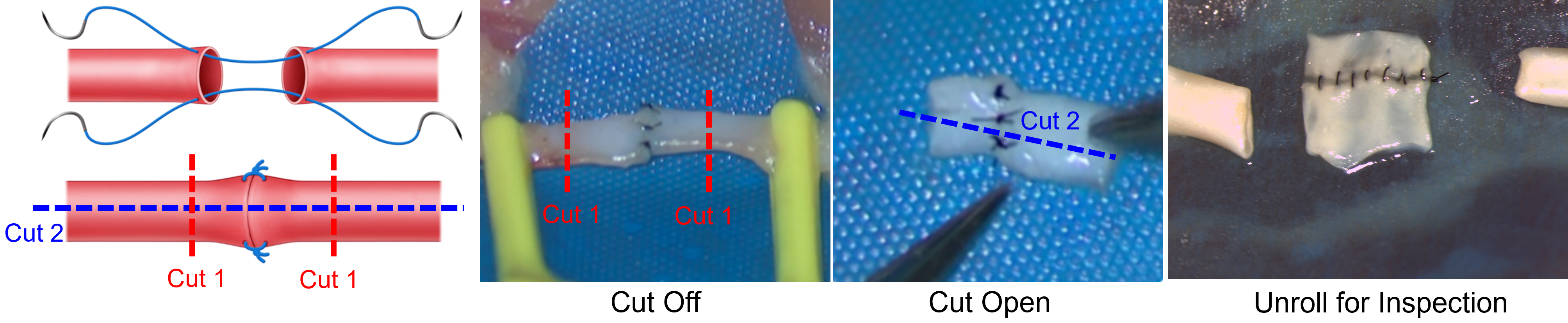} 
    \caption{The key steps after completing the anastomosis. The sutured section is cut from the two vessels, cut open, and unrolled to inspect the placement of the stitches.
    }
\label{fig:fig2-cut-and-unroll}
\end{figure}
\begin{figure}[!t]
    \centering
\includegraphics[width=0.495\textwidth]{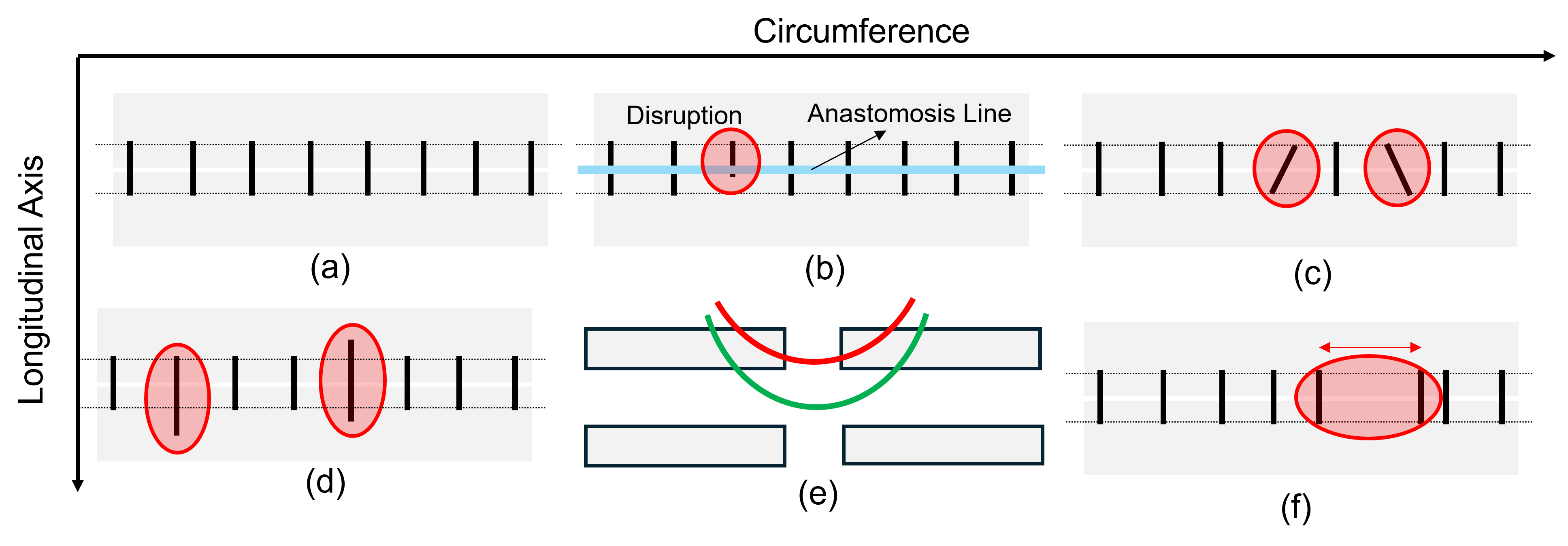} 
    \caption{Errors under unrolled view. (a) Error-free anastomosis. (b) E.1: Disruption of anastomosis line. (c) E.2:  Oblique stitch. (d) E.3: Too wide bite length. (e) E.4: Partial thickness stitch. (f) E.5: Unequal distance between stitches.
    }
\label{fig:error_sketch}
\end{figure}

Microsurgical anastomosis involves the precise surgical connection of two tiny structures, usually tubular ones like blood vessels or veins, performed at a microscopic scale. After all stitches are completed, as shown in Fig.~\ref{fig:fig2-cut-and-unroll}, the sutured vessel is cut off along the cut line 1, cut open along the cut line 2, and unrolled to inspect anastomosis quality. We quantify 5 errors in the end-product of the anastomosis. Different errors are induced by inaccurate positioning of the proximal and distal insertion points in distinct spatial dimensions.  
These errors are defined as follows. Refer to Fig.~\ref{fig:error_sketch} for conceptual depictions of the errors, Table~\ref{tab:geometric_metrics} for the mathematical definitions, and Fig.~\ref{fig:fig5-geometry} for illustrations of the defined symbols.

\begin{itemize}
\item[(E.1)] \textit{Disruption of the anastomosis line}: This error occurs when the stitches are asymmetrical relative to the anastomosis line, which is the conceptual line marking the junction where the two vessels meet. 
\item[(E.2)] \textit{Placing an oblique stitch}: This error occurs when stitches are not perpendicular to the anastomosis line. 

\item[(E.3)] \textit{Taking too wide a bite}: This error occurs when the distance between the distal and proximal points deviates from the required value of three times the needle diameter. 

\item[(E.4)] \textit{Partial thickness stitch}: This involves placing a stitch that doesn't penetrate the full thickness of the vessel, allowing the intimal layer to cover the stitch partially. 

\item[(E.5)] \textit{Unequal distance between stitches}: 
This error occurs when the spacing between stitches is inconsistent. 
\end{itemize}

The defined errors 
% (\textit{E.i} where $i\in\{1,2,\dots, 5\}$)
are critical mistakes observed in anastomosis procedures. These errors are central to the ALI score method, a widely used metric for evaluating microsurgeons' proficiency in microsurgical anastomosis \cite{ghanem2016anastomosis}. 
The ALI method is used to assess the proficiency levels of surgeons performing microsurgical interventions. It functions as a metric that allows experts to categorize trainees based on their performance. 

\section{Dataset}
\label{sec:dataset}
We used three datasets in this study (see Table~\ref{tab:datasets}). 
Our dataset was generated during experiments conducted at the Iuliu Hațieganu University of Medicine and Pharmacy in Cluj-Napoca, Romania (D.1 and D.2), and at Erasmus University Medical Center in Rotterdam, the Netherlands (D.3). 
The experimental task involved performing an end-to-end microsurgical anastomosis on a chicken leg model\footnote{Models employed in training programs range from low-fidelity options like silicone tubes to high-fidelity alternatives involving animals. Intermediate-fidelity options, such as chicken leg, are also widely used due to their reproducibility, cost-effectiveness, and ethical acceptability \cite{hino2003training}.}, with 8 stitches, each secured by two knots. 
Before the experiments began, we verbally described the task and provided information on the recording equipment and the collected data.
After that, participants signed the consent forms. 
After completing the task, they used a microscope to capture the anastomosis photos. Photos of insufficient quality were excluded, and the remaining images were anonymized before being reviewed by the experts to minimize potential bias.
The experiment and data collection were conducted following the ethical practices outlined in the EPFL institutional guidelines. The demographic information of the participants is listed in Table~\ref{tab:datasets}. 
Each dataset was recorded in different conditions, leading to images with drastically distinct backgrounds and noise. While this lack of uniformity made it more challenging to analyze data, it was also done on purpose to test the sensitivity of our pipeline in various conditions. 

\begin{table*}[!t]
\centering
\caption{Defined geometric variables and metrics used for quantifying errors.}
\label{tab:geometric_metrics}
\begin{adjustbox}{width=0.935\textwidth}
\begin{tabularx}{\textwidth}{ >{\hsize=0.09\hsize\centering\arraybackslash}X  >{\hsize=0.51\hsize\raggedright\arraybackslash}X  >{\hsize=0.4\hsize\raggedright\arraybackslash}X } % Adjusted column widths
\toprule
\textbf{Symbol} & \textbf{Explanation} & \textbf{Formula} \\ 
\hline
\multicolumn{3}{l}{\textbf{Scalars}} \\
$a^i$ & Aspect ratio of the $i$-th stitch r-bbx & $a^i = \frac{\|\mathbf{l}^i_w\|}{\|\mathbf{l}^i_h\|}$ \\
$\alpha^i$ & Stitch orientation deviation of the $i$-th stitch r-bbx & $\alpha^i = \arccos\left(\hat{\mathbf{l}}_o^i \cdot \hat{\mathbf{l}}_f\right)$ \\
$\beta^i$ & Inter-stitch curvature deviation & $\beta^i = \arccos\left(\frac{\mathbf{l}_c^i}{\|\mathbf{l}_c^i\|} \cdot \frac{\mathbf{l}_c^{i+1}}{\|\mathbf{l}_c^{i+1}\|}\right)$ \\
${{l^i_{wn}}}$ & Normalized stitch width of the $i$-th stitch & ${{l^i_{wn}}} = \frac{\|\mathbf{l}^i_w\|}{\|\mathbf{l}_{vf}\|}$ \\
$l_d^i$ & Normalized Inter-stitch distance & $l_d^i = \frac{\left(\frac{\mathbf{l}_c^i}{\|\mathbf{l}_c^i\|} \cdot \hat{\mathbf{l}}_f\right)}{\|\mathbf{l}_{vf}\|}$ \\ 
$N,n$ & Number of ground truth and detected stitches & \\
$\alpha^\star, \beta^\star, a^\star, \newline l_w^-, l_w^+,  l_d^-, l_d^+ $ & \parbox[c]{\hsize}{Identified threshold values} &  \\

\midrule
\multicolumn{3}{l}{\textbf{Vectors}} \\
$\mathbf{c}_s^i$ & Center of the $i$-th stitch r-bbx & Computed based on Mask R-CNN prediction \\
$\hat{\mathbf{l}}_o^i$ & Unit constructing vector of the $i$-th stitch r-bbx that aligns with $\hat{\mathbf{l}}_f$ & $\hat{\mathbf{l}}^i_o = \arg \max_{\mathbf{v} \in \{\frac{\mathbf{l}^i_w}{\|\mathbf{l}^i_w\|}, \frac{\mathbf{l}^i_h}{\|\mathbf{l}^i_h\|}\}} \left( \mathbf{v} \cdot \hat{\mathbf{l}}_f \right)$ \\
$\mathbf{l}^i_w$,$\mathbf{l}^i_h$ & Constructing vector of the $i$-th stitch r-bbx for width and height & Computed based on four corners of the $i$-th stitch r-bbx\\

$\mathbf{l}_c^i$ & Vector linking the centers of neighboring $\mathbf{c}_s^i$ & \\
$\hat{\mathbf{l}}_f$ & Unit vector representing the approximated anastomosis line & First principal component of the matrix formed by $\mathbf{c}_s^i$. \\
$\mathbf{l}_{vf}$ & Constructing vector of the vessel r-bbx that aligns with $\hat{\mathbf{l}}_f$ & Computed similarly to $\hat{\mathbf{l}}_o^i$ without normalization. \\ \midrule
\multicolumn{3}{l}{\textbf{Number of Errors}} \\
$s_1$ & Number of E.1 (disruption of anastomosis line) & $s_1 = \sum_{i=1}^{n - 2} \mathbf{1}_{C_1}(\beta^i)$ with $C_1 = \{x > \beta^{\star}\}$ \\
$s_2$ & Number of E.2 (oblique stitch) & $s_2 = \sum_{i=1}^{n} \mathbf{1}_{C_2}(\alpha^i)$ with $C_2 = \{x > \alpha^{\star}\}$ \\
$s_3$ & Number of E.3 (too wide a bite) & $s_3 = \sum_{i=1}^{n} \mathbf{1}_{C_3}({{l^i_{wn}}})$ with $C_3 = \{x > {l_{w}}^{+}\}$ \\
$s_4$ & Number of E.4 (partial thickness stitch) & $s_4 = \sum_{i=1}^{n} \mathbf{1}_{C_4}(a^i) + \sum_{i=1}^{n} \mathbf{1}_{C_5}({{l^i_{wn}}}) + N - n$ with $C_4 = \{x < {a}^{\star}\}$ and $C_5 = \{x < {l_w}^{-}\}$ \\
$s_5$ & Number of E.5 (unequal distancing) & $s_5 = \sum_{i=1}^{n} \mathbf{1}_{C_5}(\mathbf{l}_d^i)$ with $C_5 = \{l_d^- < x < l_d^+\}$ \\
\bottomrule
\end{tabularx}
\end{adjustbox}
\end{table*}

\begin{table}[!t]
\centering
\caption{The demographic information of the participants.}
\begin{adjustbox}{width=0.45\textwidth}
\begin{threeparttable}
\begin{tabular}{lcccccccc}
\toprule
 & \multicolumn{3}{c}{\textit{Datasets}}  & \multicolumn{5}{c}{\textit{Demographics}} \\ \midrule
{\textbf{Expertise level}} & \textbf{D.1} & \textbf{D.2} & \textbf{D.3} & \begin{tabular}[c]{@{}c@{}}\textbf{Age}\\ {[}\textbf{Years}{]}\end{tabular} & \multicolumn{2}{c}{\textbf{Gender}\tnote{1}} & \multicolumn{2}{c}{\textbf{Handedness}\tnote{2}} \\ \midrule

{Surgeon} & 3 & 1 &9\tnote{*} & $32.3 \pm 2.3$  & 3 F  & 6 M   & 1 L  & 8 R \\
{Resident} & 10 & 7 & 0 & $26.5 \pm 2.6$ & 9 F  & 8 M  & 0 L  & 17 R\\
{Student} & 10 & 10  & 0 & $22.7\pm 1.6$ & 13 F  & 7 M  & 1 L & 19 R\\ \midrule
{Images} & 23  & 18  & 9  & \multicolumn{5}{c}{-} \\ 
\bottomrule
\end{tabular}
  \begin{tablenotes}
 \item[1] Biological genders: `M' and `F' represent males and females, respectively.
  \item[2] `R' and `L' indicate right and left, respectively. Handedness is determined using the questionnaire introduced in \cite{oldfield1971assessment}.
  \item[*] 4 participants completed two trials each, while one participant completed a single trial.
  \end{tablenotes}
  \end{threeparttable}
\end{adjustbox}
\label{tab:datasets}
\end{table}

\section{Image Processing Framework for Assessing Microsurgical Anastomosis}
\label{section:framework}
The block diagram of the proposed framework for assessing microsurgical skills is shown in Fig.~\ref{fig:fig1}. 

\subsection{Segmentation Pipeline}
\label{section:segmentation-pipeline}
We use mask region-based convolutional neural network (Mask R-CNN) \cite{he2017mask} with a ResNet50-FPN backbone to detect and segment vessel and suture instances within the anastomosis images provided in the dataset, as illustrated in Fig.~\ref{fig:fig1}.
For every detected instance in the input image, the network provides (1) the coordinates of the predicted bounding box encapsulating the instance, (2) probabilities associated with the relevant class labels for the instance, and (3) a probability mask delineating the pixel-level segmentation of the instance.
We apply a class threshold of $0.5$ to filter out all detected instances, and a mask threshold of $0.5$ to produce a binary segmentation mask for each instance.

Using OpenCV, we extract the contour points from a binary mask and then fit a minimum area rectangle around those points \cite{suzuki1985topological}.
While the predicted bounding box by the employed Mask R-CNN provides position and scale information of the detected instances, the fitted minimum-area rectangle additionally captures the orientation data that is crucial for robust error detection.
We refer to the estimated minimum-area rectangle as a rotated bounding box (r-bbx) and leverage its geometric properties to calculate the geometric metrics used for anastomosis evaluation.

\subsection{Geometric Representation} 
\label{subsection:geometry}
\begin{figure}[!t]
    \centering
    \includegraphics[width=0.375\textwidth]{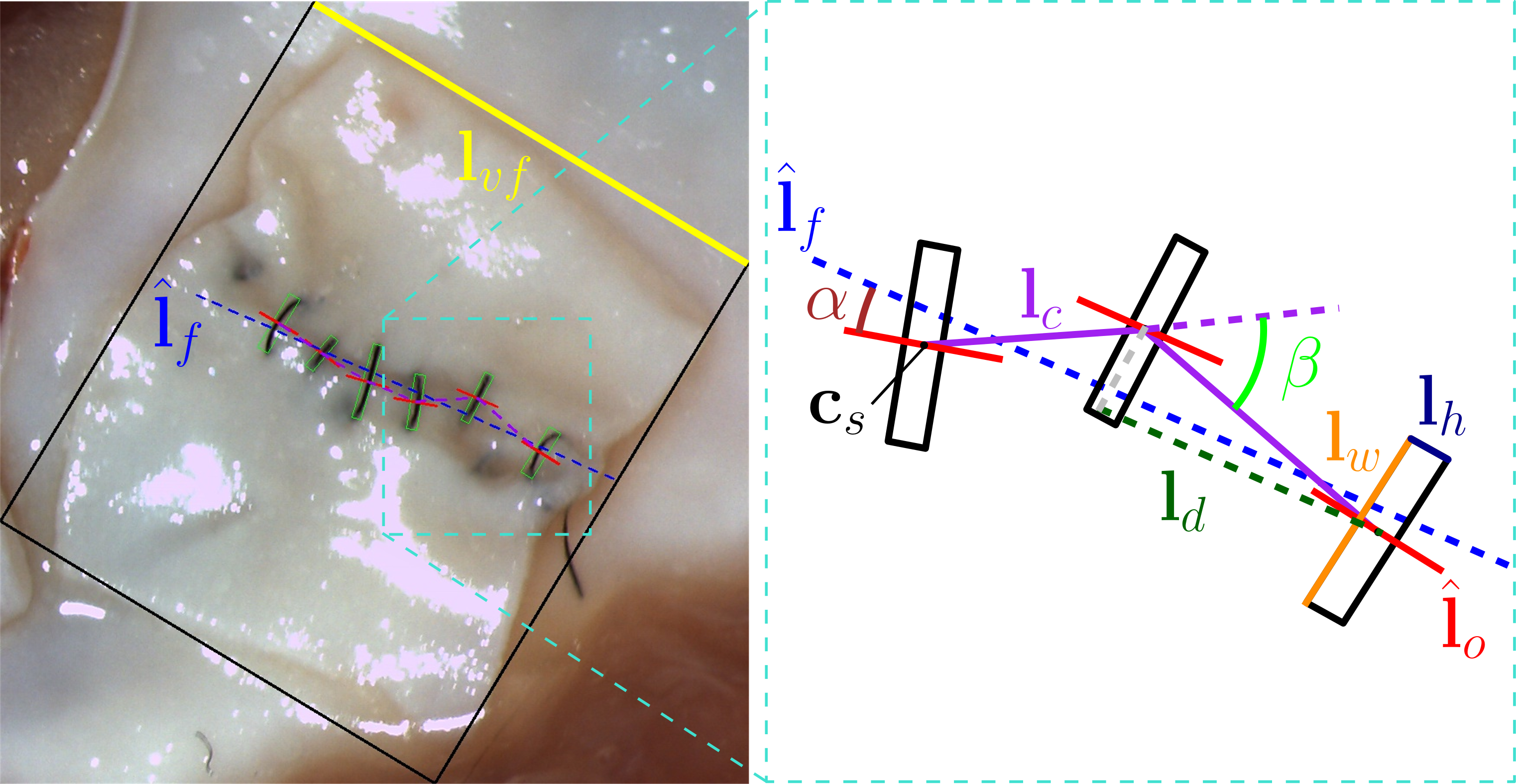} 
    \caption{Anastomosis line, stitches' bounding boxes, and inter-stitches relationships, geometric attributes used for automatic scoring: (left) example of extracted geometric attributes superimposed to the anastomosis image, and (right) annotation of the attributes on a zoomed-in area of three stitches.}
    \label{fig:fig5-geometry}
\end{figure}

To automate the scoring, we developed a geometric representation of the expected anastomosis. As shown in Fig. \ref{fig:fig5-geometry}, it encompasses three main attributes: the anastomosis line, the width, and height of each stitch (the r-bbx characteristics), and the inter-stitches geometric relationships. Refer to Table~\ref{tab:geometric_metrics} for the mathematical definitions. We compare relevant geometric attributes against predefined threshold values to detect error stitches, as detailed in the final section of Table~\ref{tab:geometric_metrics}.

We introduce a general algorithm for identifying threshold values based on expert surgeons' evaluation. For an anastomosis image, multiple stitches are evaluated by human raters to detect potential errors. However, human raters provide only a binary indication of error presence or a discrete count of total errors, without specifying which stitch is erroneous. To infer which stitches are identified as erroneous and to determine the criteria used by humans to judge these errors, we formulate the error quantification as a multiple-instance classification problem as follows:

\noindent
\textit{Classification}: Given a set of geometric attributes of the stitches $\mathcal{K} = \{\mathbf{k}^i\}$, where $i \in \{1,\, \dots,\, k\}$ and $k \leq n$, we apply a threshold-based binary classifier to each element of $\mathcal{K}$. 
If an attribute meets a defined threshold condition $C$, we label it as True, indicating an abnormal attribute.

\noindent
\textit{Identification}: We incorporate the human score to determine the classifier's threshold value. If the human score is $s_h$, we select $s_h$ attributes from $\mathcal{K}$ that are most likely to meet the condition $C$ of the classifier. For example, if the condition is $C = \{x > k^\star\}$ and $s_h = 3$, we select three attributes with the highest values; if the condition is $C = \{x < k^\star\}$, we select the 3 attributes with the lowest values. These selected attributes are labeled as True. If $s_h = 0$, we select one attribute from $\mathcal{K}$ that is most likely to satisfy the condition and label it as False. Finally, for all participants in the dataset, we group the selected attributes and their corresponding labels to form a dataset and compute the receiver operating characteristic (ROC) curve for a binary classifier model. The threshold for the error detection algorithm is then chosen to optimize the difference between the true positive rate and the false positive rate. If the classifier uses multiple thresholds, we employ a grid search algorithm, allowing only one threshold value to be identified by the ROC curve to search through the potential space of threshold values. In the end, the combination of threshold values that maximizes the difference between the true positive rate and the false positive rate is selected.

\subsection{Fine-tuning on Custom Dataset}
The model is initialized with weights pre-trained on the COCO dataset \cite{lin2014microsoft} and then fine-tuned on our custom dataset.
This dataset consists of 50 high-quality images selected from our dataset. We included only those images that met our quality standards for analysis while discarding any poorly captured images. 
The appearance of stitches and vessels varies across the three training sets. To address potential bias from imbalanced image counts in the different datasets, we employed a balanced sampling strategy. 
In each training iteration, we randomly select a sub-dataset with equal probability and then sample an image from the chosen sub-dataset to include in the training batch. This method ensures that each sub-dataset contributes equally to the training process, maintaining a balanced data representation and preventing any single session from disproportionately affecting the model's learning.
To address the limited amount of available data and enhance the model's robustness, we apply image augmentation techniques based on the principles outlined in \cite{muller2021trivialaugment}. This involves adjusting parameters such as scale, orientation, color, and contrast to effectively augment the input images.

For training, we employ the Adam optimizer with an accumulated batch size of 8 and an initial learning rate of $5 \times 10^{-4}$. To enhance the learning process, we use a linear warm-up scheduler for the first 10 epochs, followed by an exponential decay scheduler for the remaining 190 epochs.
During the training phase, we monitor the model's performance on a validation set and save the model weights that achieve the highest values for the considered metrics for bounding boxes and masks.
These metrics, which will be detailed in Section~\ref{subsection:detection}, are essential for evaluating the model's performance. The saved model weights are subsequently used to assess the model on the test set.
In this work, we evaluate the model's performance using metrics based on \textit{average precision} ($\text{AP}$), a widely used evaluation metric for object detection and segmentation that accounts for both precision and recall \cite{carion2020end}. 

To thoroughly evaluate the model’s performance in detecting and segmenting vessels and stitches in unseen images, we perform a 5-fold cross-validation on the custom dataset.
Images from each dataset are independently divided into 5 mutually exclusive folds. In each validation iteration, one fold from each session is selected and combined to create the test set, while the remaining folds from each session are merged and randomly split into training and validation sets with a 75$/$25 ratio. 
We report {$\text{AP}_b$} and {$\text{AP}_m$} metrics. 
The calculation of {$\text{AP}^{50}_b$} involves matching predictions with ground truth at an IoU threshold of $0.5$, computing precision and recall, and calculating the area under the precision-recall curve to evaluate detection performance. 
This process is repeated across multiple IoU thresholds and averaged to obtain the overall {$\text{AP}_b$} and {$\text{AP}_m$}, which provides a comprehensive measure of the model's performance in detecting and segmenting objects across different levels of overlap with the ground truth.

\section{Data Analysis}
\label{section:data-analysis}
\subsection{Reliability of the Anastomosis Lapse Index (ALI)}
\label{subsection:reliability}
We performed an intra-rater reliability analysis to examine the impact of time on a single expert's judgments. A microsurgeon evaluated 23 out of 31 anastomosis photos from D.1 twice, with a six-month interval between evaluations. This time frame aligns with the typical duration of a microsurgery training cycle, providing a realistic assessment of the long-term consistency of human ratings. 
The intra-rater reliability is quantified with two levels of accuracy: true accuracy and binary accuracy. True accuracy is calculated by summing true positives and true negatives and dividing by the total number of evaluated photos. Binary accuracy, on the other hand, employs a more relaxed criterion for true positives. 
To compute binary accuracy, the test and ground truth scores are converted into binary values before accuracy is determined. 

\begin{figure}[!t]
\centering
\includegraphics[width=0.485\textwidth]{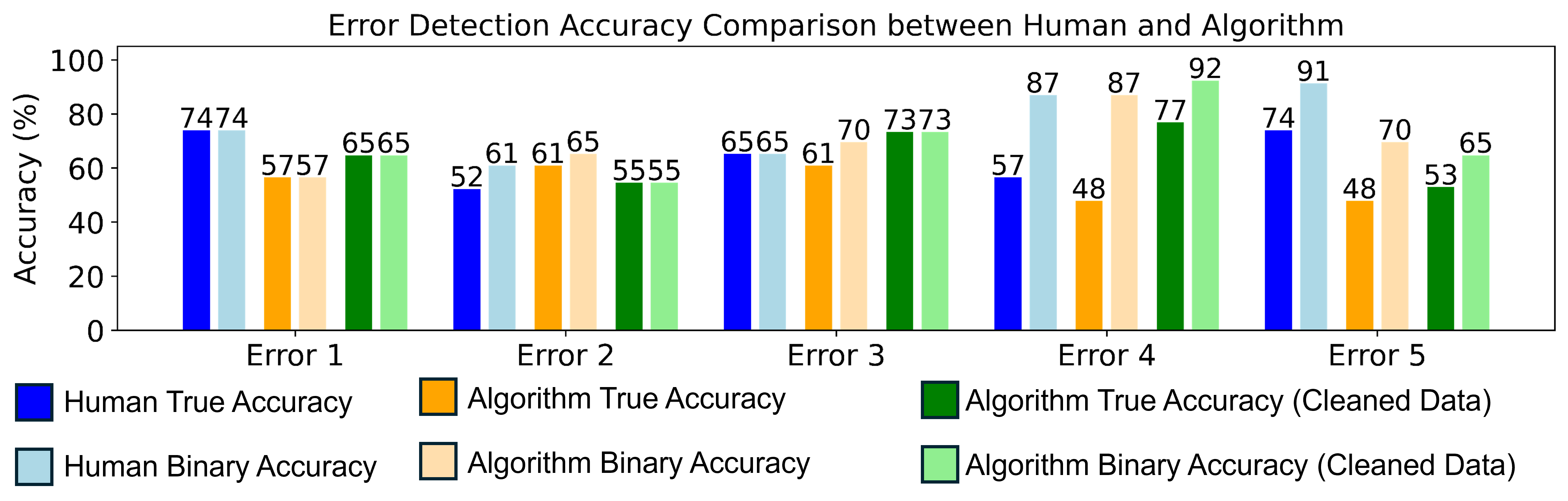}
\caption{Comparison of error detection accuracy between the human rater and our algorithm. 
Human accuracy is assessed by evaluating the consistency between two trials separated by a six-month interval (intra-rater reliability analysis).
}
\label{fig:fig7-performance}
\end{figure}

The accuracy results for evaluating human intra-reliability are presented in Fig.~\ref{fig:fig7-performance}. Human true and binary accuracy in detecting E.2 (i.e., placing an oblique stitch) are notably lower compared to other errors. This discrepancy may suggest a shift in obliqueness error detection criteria over time or underscore the inherent difficulty in accurately estimating stitch rotation. 
In our dataset, the vessels are positioned in arbitrary orientations. The ability to consistently estimate an object's orientation when the reference frame is shifted—a visual-spatial ability known as field dependence \cite{mcgee1979human}—is influenced by the degree of rotation of the reference frame \cite{corbett2006observer}.
Furthermore, we observe a noticeable difference between human true  and binary accuracy in detecting E.4 (partial thickness stitch) and E.5 (unequal distancing between stitches). \textit{This disparity suggests that certain errors are more identifiable, while others are prone to causing confusion}.

\subsection{Accuracy at Error Detection as Compared to Humans'}
\label{subsection:human-algorithm-error-detection}
We assess the performance of our proposed error detection system by comparing it to the assessments made by the human raters.
We evaluate the algorithm's true and binary accuracy under two conditions. In the first condition, the algorithm's thresholds are set using the original data according to the procedure detailed in Section~\ref{subsection:geometry}. This condition assesses the algorithm's performance when trained on potentially noisy labeled data. In the second condition,  we refine the original data by excluding images with conflicting scores between the two human ratings used for intra-reliability analysis (``Cleaned Data''). Note that the number of inconsistently scored images varies across different errors. Ultimately, 6, 12, 8, 10, and 6 inconsistent ratings were removed from the ratings to identify thresholds for E.1 (anastomosis line disruption), E.2 (oblique stitch), E.3 (too wide bite), E.4 (partial thickness stitch), and E.5 (unequal distance between stitches). 
After cleaning the dataset, the algorithm’s thresholds are recalculated, and its accuracy is reassessed. This evaluation determines how well the algorithm aligns with the human rater’s internal error detection model.

The algorithm's accuracy based on these two datasets is illustrated in Fig. \ref{fig:fig7-performance}, while the thresholds identified using these datasets are reported in Table~\ref{table:threshold_comparison}. We found a slight difference between the algorithm thresholds determined using the original and cleaned dataset. The algorithm's accuracy—both true and binary—surpasses human accuracy for errors E.3 and E.4 but is either lower than or comparable to human accuracy for detecting errors E.1, E.2, and E.5. 
When algorithms outperform humans in accuracy, it often highlights their effectiveness, especially in areas like image-based disease diagnostics \cite{Liu2019A}. Such tasks typically depend on the model's ability to identify complex patterns that are crucial for accurate error detection.
However, in this study, the errors pertain to simpler geometric metrics like stitch width, orientation, and inter-stitch distance, as detailed in the ALI score description. \textit{The introduced computer-based method is well-suited to provide consistent and precise measurements of these metrics}, facilitating reliable error detection. If the algorithm demonstrates lower accuracy when human labeling is used as the ground truth, it might suggest potential issues with the human scoring process, thereby offering valuable feedback for improving human assessment.

\begin{table}[!t]
\centering
\caption{Comparison of identified algorithm thresholds based on human trial 2 scores and a cleaned version of human trial 2 scores. }
\begin{adjustbox}{width=0.39\textwidth}
\begin{tabular}{lccccccc}
\toprule
 & \multicolumn{7}{c}{\textbf{Threshold}} \\
\cmidrule(lr){2-8}
 \textbf{Dataset} & $\beta^\star$ & $\alpha^\star$ & $l_w^+$ & $a^\star$ & $l_w^-$ & $l_d^-$ & $l_d^+$ \\
\midrule
T.2 & 29.80 & 38.11 & 0.13 & 2.43 & 0.06 & 0.07 & 0.148 \\
Cleaned T.2 & 29.80 & 38.11 & 0.16 & 2.43 & 0.06 & 0.07 & 0.149 \\
\bottomrule
\end{tabular}
\end{adjustbox}
\label{table:threshold_comparison}
\end{table}

\begin{table}[!t]
\caption{Metrics to analyze the model's performance across the \textit{Stitch} and \textit{Vessel} classes in the three datasets. The highest performance across the datasets is highlighted.}
\centering
\begin{adjustbox}{width=0.325\textwidth}
\begin{tabular}{lccccc}
\toprule
 & & \multicolumn{4}{c}{\textbf{Metric}} \\
\cmidrule(lr){3-6}
\textbf{Class} & \textbf{Dataset}  & $\textbf{AP}_b$ & \textbf{$\text{AP}^{50}_b$} & \textbf{$\text{AP}_{{m}}$} & \textbf{$\text{AP}_{{m}}^{50}$} \\
\midrule
\textit{Stitch} & All & 0.877 & 0.891 & 0.319 & 0.819 \\ 
\textit{Vessel} & All & 0.760 & 0.974 & 0.918 & 0.974 \\ 

\midrule
\multirow{3}{*}{\textit{Stitch}} & D.1 & \cellcolor{lightgray}0.945 & \cellcolor{lightgray}0.952 & \cellcolor{lightgray}0.413 & \cellcolor{lightgray}0.934 \\ 
 & D.2 & 0.798 & 0.829 & 0.231 & 0.671 \\ 
 & D.3 & 0.898 & 0.898 & 0.296 & 0.822 \\ 
\midrule
\multirow{3}{*}{\textit{Vessel}} & D.1 & 0.748 & 0.950 & 0.875 & 0.950 \\
 & D.2 & 0.772 & \cellcolor{lightgray}1.000 & \cellcolor{lightgray}0.990 & \cellcolor{lightgray}1.000 \\ 
 & D.3 & \cellcolor{lightgray}0.865 & \cellcolor{lightgray}1.000 & 0.940 & \cellcolor{lightgray}1.000 \\  
\bottomrule
\end{tabular}
\end{adjustbox}
\label{table:metrics}
\end{table}

\begin{figure*}[!t]
    \centering
    \includegraphics[width=0.90\textwidth]{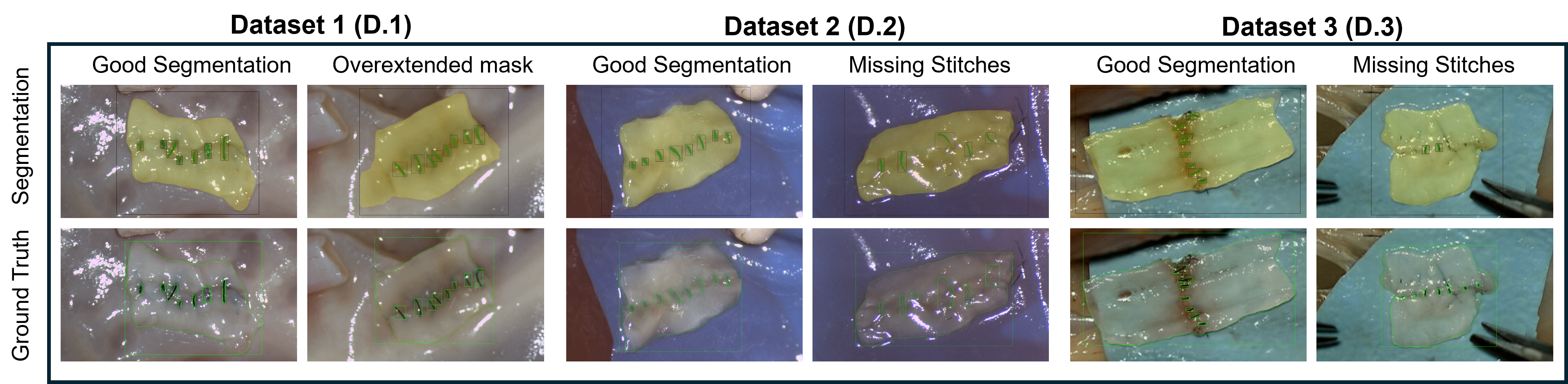} 
    \caption{Examples of segmentation quality across datasets. The first row displays segmentation results: stitch instances are marked in green, vessel instances in yellow, and bounding boxes in black. The second row shows the ground truth annotations. In D.1, segmentation mismatches are observed, often due to low contrast between vessels and their backgrounds. In D.2 and D.3, the error of missing stitches is more evident, potentially resulting from subtle thread colors and low image resolution.}
    \label{fig:fig10-detection-results}
\end{figure*}

\subsection{Detection and Segmentation Performance}
\label{subsection:detection}
The evaluation metrics presented in Table~\ref{table:metrics} provide a thorough analysis of the model's performance across the Stitch and Vessel classes, evaluated on three distinct datasets introduced in Section~\ref{sec:dataset}.
For the Stitch class, the results reveal varying performance across the datasets. D.1 demonstrates the highest performance of $\text{AP}_b = 0.945$ and $\text{AP}_m = 0.413$, likely due to higher image quality (black thread and high resolution) or lower variability within the dataset. In contrast, D.2 shows the lowest performance, which we attribute to poor data quality, specifically the blue thread blending with the vessel background and making it difficult to distinguish. D.3 shows average performance. 
This moderate outcome is partly due to the use of black thread; however, the lower resolution of this dataset, compared to others, poses challenges in detecting small objects such as stitches. 
We also notice the discrepancy between $\text{AP}_m$ and $\text{AP}_m^{50}$, which indicates although the stitch instances can be detected reliably, the performance of mask segmentation at higher IoU thresholds remains inadequate. 
As our pipeline relies on the network’s mask output to extract geometric information, improving this accuracy is essential for future work. This improvement could be achieved by augmenting the training dataset and adopting network architectures that prioritize precise segmentation \cite{wang2020deep}. For the Vessel class, the model achieves higher $\text{AP}_m$ compared to the Stitch class across all datasets, despite the number of vessel instances being much lower. Among the datasets, D.2 achieves the highest performance, with $\text{AP}_m = 0.99$, likely attributed to the vessels in this dataset being displayed against a blue background, which enhances their visibility and facilitates more accurate segmentation. In contrast, in D.1, the vessels are often placed against a white background. This background, being similar in color to the vessels, complicates the task of precise segmentation. Example images demonstrating the detection and segmentation results are included in Fig.~\ref{fig:fig10-detection-results}.

\section{Concluding Remarks}
\label{sec:discussion}
Systematic and quantitative approaches to assessing anastomosis photos in microsurgical interventions are essential, as current methods are primarily subjective and rely on questionnaires. Furthermore, research in this area is limited, underscoring the need for greater attention and investigation. Existing similar studies either focus on different surgical practices (e.g., \cite{frischknecht2013objective} in open surgery), employ low-fidelity models \cite{noraset2024automated}, or lack full automation \cite{kim2020end}. Leveraging image processing techniques and the extraction of geometric features from vessels and stitches, we developed an automated framework for quantitative assessment of microsurgical anastomosis images. This framework not only provides error analysis and feedback to trainees but also differentiates between various levels of expertise among them.

\begin{figure*}[!t]
    \centering
    \includegraphics[width=0.825\textwidth]{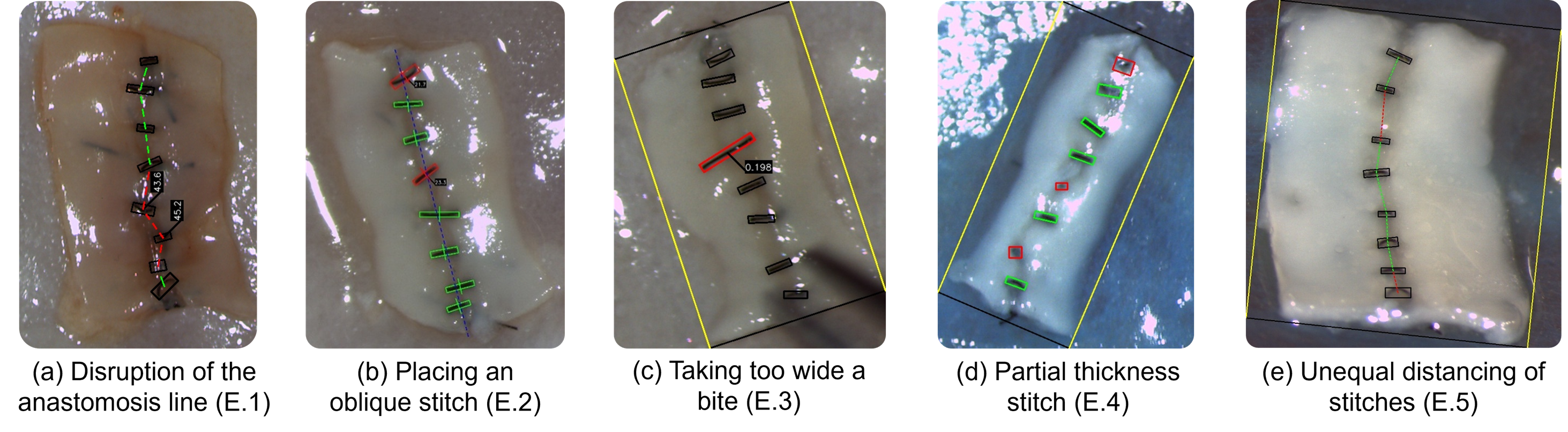} 
    \caption{Examples of errors detected by our framework, with errors highlighted in red and regular attributes marked in green. }
    \label{fig:fig10-errors-visualization}
\end{figure*}

A key feature of our research is its focus on error modeling and quantification based on the errors defined by the ALI score method. Unlike previous studies that directly extract geometric attributes of stitches to assess expertise levels \cite{noraset2024automated}, or as in \cite{kim2020end}, where the statistics of these attributes are leveraged to develop a new scoring system, our framework establishes a precise mapping between the specific errors outlined by the ALI score and the used geometric attributes. Employing this approach facilitates the straightforward integration of our automatic identification system with existing training programs, minimizing the need for extra educator and student retraining and administrative efforts typically required when adopting new tools in surgical education.

Conventional ALI score assessments by human raters typically provide only the cumulative number of errors in the final image, leading to potential ambiguities and a lack of specific feedback. Similarly, while other computer-assisted approaches in the literature analyzed geometric attributes statistically \cite{solis2008development,frischknecht2013objective,kim2020end}, they do not pinpoint which specific stitches contribute to these errors. Our proposed quantitative framework enables detailed analysis by detecting and analyzing errors on a per-stitch basis. As demonstrated in Fig. \ref{fig:fig10-errors-visualization}, errors are highlighted in red, while regular attributes are marked in green. This granular feedback is vital for training purposes. Additionally, this detailed visualization of geometric attributes may allow human raters to perform more consistent evaluations, thereby improving the reliability of the assessment process.

\subsection{Limitations and Future Work}
\label{subsec:limitations}
Currently, and to the best of our knowledge, datasets related to microsurgical anastomosis photos used in the literature are not publicly available, making it challenging to expand our dataset. In this study, our dataset includes 50 images with diverse appearances generated from three different recording sessions. However, to improve modeling and testing, future research will focus on increasing the sample size to enhance the generalizability and robustness of the proposed framework. 

Human experts assess microsurgical anastomosis errors by evaluating both the geometric attributes of stitches and the visual appearance of the vessel, focusing on issues such as tissue distortion and strangulation.
The geometric attributes addressed in this paper do not cover all ten ALI error types needed for a comprehensive assessment. For this purpose, additional visual information must be integrated into the current framework. This is feasible with sufficient training data.
Given the scarcity of data, our future studies will focus on developing a user-friendly annotation platform. Designed with medical experts in mind, this platform will facilitate the precise localization of errors within images. The additional data collected through this platform can be seamlessly integrated into our existing pipeline, enabling a more comprehensive evaluation of all ALI errors.

\bibliography{sn-bibliography}

\begin{thebibliography}{10}

\bibitem{selber2012tracking}
J.~C. Selber, E.~I. Chang, J.~Liu, H.~Suami, D.~M. Adelman, P.~Garvey, M.~M. Hanasono, and C.~E. Butler, ``Tracking the learning curve in microsurgical skill acquisition,'' {\em Plastic and reconstructive surgery}, vol.~130, no.~4, p.~550e, 2012.

\bibitem{martin1997objective}
J.~Martin, G.~Regehr, R.~Reznick, H.~Macrae, J.~Murnaghan, C.~Hutchison, and M.~Brown, ``Objective structured assessment of technical skill {(OSATS)} for surgical residents,'' {\em British Journal of Surgery}, vol.~84, no.~2, pp.~273--278, 1997.

\bibitem{gofton2012ottawa}
W.~T. Gofton, N.~L. Dudek, T.~J. Wood, F.~Balaa, and S.~J. Hamstra, ``The ottawa surgical competency operating room evaluation {(O-SCORE)}: a tool to assess surgical competence,'' {\em Academic Medicine}, vol.~87, no.~10, pp.~1401--1407, 2012.

\bibitem{satterwhite2014stanford}
T.~Satterwhite, J.~Son, J.~Carey, A.~Echo, T.~Spurling, J.~Paro, G.~Gurtner, J.~Chang, and G.~K. Lee, ``The stanford microsurgery and resident training {({SMaRT})} scale: validation of an on-line global rating scale for technical assessment,'' {\em Annals of plastic surgery}, vol.~72, pp.~S84--S88, 2014.

\bibitem{ghanem2016anastomosis}
A.~M. Ghanem, Y.~Al~Omran, B.~Shatta, E.~Kim, and S.~Myers, ``Anastomosis lapse index ({ALI}): a validated end product assessment tool for simulation microsurgery training,'' {\em Journal of Reconstructive Microsurgery}, vol.~32, no.~03, pp.~233--241, 2016.

\bibitem{moorthy2003objective}
K.~Moorthy, Y.~Munz, S.~K. Sarker, and A.~Darzi, ``Objective assessment of technical skills in surgery,'' {\em Bmj}, vol.~327, no.~7422, pp.~1032--1037, 2003.

\bibitem{ebina2021motion}
K.~Ebina, T.~Abe, M.~Higuchi, J.~Furumido, N.~Iwahara, M.~Kon, K.~Hotta, S.~Komizunai, Y.~Kurashima, H.~Kikuchi, {\em et~al.}, ``Motion analysis for better understanding of psychomotor skills in laparoscopy: objective assessment-based simulation training using animal organs,'' {\em Surgical endoscopy}, vol.~35, pp.~4399--4416, 2021.

\bibitem{gholami2024objective}
S.~Gholami, A.~Manon, K.~Yao, A.~Billard, and T.~R. Meling, ``An objective skill assessment framework for microsurgical anastomosis based on {ALI} scores,'' {\em Acta Neurochirurgica}, vol.~166, no.~1, pp.~1--13, 2024.

\bibitem{kim2020end}
E.~Kim, I.~C. Norman, S.~Myers, M.~Singh, and A.~Ghanem, ``The end game--a quantitative assessment tool for anastomosis in simulated microsurgery,'' {\em Journal of Plastic, Reconstructive \& Aesthetic Surgery}, vol.~73, no.~6, pp.~1116--1121, 2020.

\bibitem{noraset2024automated}
T.~Noraset, P.~Mahawithitwong, W.~Dumronggittigule, P.~Pisarnturakit, C.~Iramaneerat, C.~Ruansetakit, I.~Chaikangwan, N.~Poungjantaradej, and N.~Yodrabum, ``Automated measurement extraction for assessing simple suture quality in medical education,'' {\em Expert Systems with Applications}, vol.~241, p.~122722, 2024.

\bibitem{hino2003training}
A.~Hino, ``Training in microvascular surgery using a chicken wing artery,'' {\em Neurosurgery}, vol.~52, no.~6, pp.~1495--1498, 2003.

\bibitem{oldfield1971assessment}
R.~C. Oldfield, ``The assessment and analysis of handedness: the edinburgh inventory,'' {\em Neuropsychologia}, vol.~9, no.~1, pp.~97--113, 1971.

\bibitem{he2017mask}
K.~He, G.~Gkioxari, P.~Doll{\'a}r, and R.~Girshick, ``Mask r-cnn,'' in {\em Proceedings of the IEEE international conference on computer vision}, pp.~2961--2969, 2017.

\bibitem{suzuki1985topological}
S.~Suzuki {\em et~al.}, ``Topological structural analysis of digitized binary images by border following,'' {\em Computer vision, graphics, and image processing}, vol.~30, no.~1, pp.~32--46, 1985.

\bibitem{lin2014microsoft}
T.-Y. Lin, M.~Maire, S.~Belongie, J.~Hays, P.~Perona, D.~Ramanan, P.~Doll{\'a}r, and C.~L. Zitnick, ``Microsoft coco: Common objects in context,'' in {\em Computer Vision--ECCV 2014: 13th European Conference, Zurich, Switzerland, September 6-12, 2014, Proceedings, Part V 13}, pp.~740--755, Springer, 2014.

\bibitem{muller2021trivialaugment}
S.~G. M{\"u}ller and F.~Hutter, ``Trivialaugment: Tuning-free yet state-of-the-art data augmentation,'' in {\em Proceedings of the IEEE/CVF international conference on computer vision}, pp.~774--782, 2021.

\bibitem{carion2020end}
N.~Carion, F.~Massa, G.~Synnaeve, N.~Usunier, A.~Kirillov, and S.~Zagoruyko, ``End-to-end object detection with transformers,'' in {\em European conference on computer vision}, pp.~213--229, Springer, 2020.

\bibitem{mcgee1979human}
M.~G. McGee, ``Human spatial abilities: psychometric studies and environmental, genetic, hormonal, and neurological influences.,'' {\em Psychological bulletin}, vol.~86, no.~5, p.~889, 1979.

\bibitem{corbett2006observer}
J.~E. Corbett and J.~T. Enns, ``Observer pitch and roll influence: the rod and frame illusion,'' {\em Psychonomic bulletin \& review}, vol.~13, no.~1, pp.~160--165, 2006.

\bibitem{Liu2019A}
X.~Liu, L.~Faes, A.~Kale, S.~Wagner, D.~J. Fu, A.~Bruynseels, T.~Mahendiran, G.~Moraes, M.~Shamdas, C.~Kern, J.~Ledsam, M.~Schmid, K.~Balaskas, E.~Topol, L.~Bachmann, P.~Keane, and A.~Denniston, ``A comparison of deep learning performance against health-care professionals in detecting diseases from medical imaging: a systematic review and meta-analysis.,'' {\em The Lancet. Digital health}, vol.~1 6, pp.~e271--e297, 2019.

\bibitem{wang2020deep}
J.~Wang, K.~Sun, T.~Cheng, B.~Jiang, C.~Deng, Y.~Zhao, D.~Liu, Y.~Mu, M.~Tan, X.~Wang, {\em et~al.}, ``Deep high-resolution representation learning for visual recognition,'' {\em IEEE transactions on pattern analysis and machine intelligence}, vol.~43, no.~10, pp.~3349--3364, 2020.

\bibitem{frischknecht2013objective}
A.~C. Frischknecht, S.~J. Kasten, S.~J. Hamstra, N.~C. Perkins, R.~B. Gillespie, T.~J. Armstrong, and R.~M. Minter, ``The objective assessment of experts’ and novices’ suturing skills using an image analysis program,'' {\em Academic Medicine}, vol.~88, no.~2, pp.~260--264, 2013.

\bibitem{solis2008development}
J.~Solis, N.~Oshima, H.~Ishii, N.~Matsuoka, K.~Hatake, and A.~Takanishi, ``Development of a sensor system towards the acquisition of quantitative information of the training progress of surgical skills,'' in {\em 2008 2nd IEEE RAS \& EMBS International Conference on Biomedical Robotics and Biomechatronics}, pp.~959--964, IEEE, 2008.

\end{thebibliography}

%%%%%%%%%%%%%%%%%%%%%%%%%%%%%%%%%%%%%%%%%%%%%%%%%%%%%%%%%%%%%%%%%%%%%%%%%%%%%%%%

\addtolength{\textheight}{-12cm}   % This command serves to balance the column lengths

%%%%%%%%%%%%%%%%%%%%%%%%%%%%%%%%%%%%%%%%%%%%%%%%%%%%%%%%%%%%%%%%%%%%%%%%%%%%%%%%

%%%%%%%%%%%%%%%%%%%%%%%%%%%%%%%%%%%%%%%%%%%%%%%%%%%%%%%%%%%%%%%%%%%%%%%%%%%%%%%%

%%%%%%%%%%%%%%%%%%%%%%%%%%%%%%%%%%%%%%%%%%%%%%%%%%%%%%%%%%%%%%%%%%%%%%%%%%%%%%%%

\end{document}